\documentclass[conference]{IEEEtran}
\IEEEoverridecommandlockouts
\usepackage{cite}
\usepackage{url}
\usepackage{amsmath,amssymb,amsfonts}
\usepackage{algorithmic}
\usepackage{graphicx}
\usepackage{textcomp}
\usepackage{amsmath}
\usepackage{xcolor}
\usepackage{multirow}
\usepackage{booktabs}
\def\BibTeX{{\rm B\kern-.05em{\sc i\kern-.025em b}\kern-.08em
    T\kern-.1667em\lower.7ex\hbox{E}\kern-.125emX}}
\begin{document}

\title{ROI-Packing: Efficient Region-Based Compression for Machine Vision}

\author{%
Md Eimran Hossain Eimon, Alena Krause, Ashan Perera, Juan Merlos, \\ Hari Kalva, Velibor Adzic, and Borko Furht\\[0.5em]
{\small
\begin{minipage}{\linewidth}
\begin{center}
Florida Atlantic University \\
\url{{meimon2021, akrause2017, aperera2016, jmerlosjr2017, hkalva, vadzic, bfurht}@fau.edu} 
\end{center}
\end{minipage}
}
}

\maketitle

\begin{abstract}
This paper introduces ROI-Packing, an efficient image compression method tailored specifically for machine vision. By prioritizing regions of interest (ROI) critical to end-task accuracy and packing them efficiently while discarding less relevant data, ROI-Packing achieves significant compression efficiency without requiring retraining or fine-tuning of end-task models. Comprehensive evaluations across five datasets and two popular end-tasks---object detection and instance segmentation---demonstrate up to a 44.10\% reduction in bitrate without compromising end-task accuracy, along with an 8.88\% improvement in accuracy at the same bitrate compared to the state-of-the-art Versatile Video Coding (VVC) codec standardized by the Moving Picture Experts Group (MPEG).

\end{abstract}


\section{Introduction}
\label{sec:intro}

Traditionally, compression techniques have been optimized for visual quality tailored to human consumption, leveraging the characteristics of the human visual system (HVS) to balance quality and efficiency. For instance, Joint Photographic Experts Group (JPEG) employs pre-defined quantization table in JPEG standard for images to discard frequency components less noticeable to human vision~\cite{iso_jpeg}, achieving compression while maintaining perceptual quality. However, the rise of artificial intelligence (AI) and the growth of machine-to-machine (M2M) communication have shifted the focus of compression from human-centric optimization to machine-oriented processing. As automated systems become increasingly prevalent, where content is analyzed and acted upon without human intervention, developing compression techniques tailored for machine vision tasks has become an urgent and critical research area. 

Compression for machine vision does not prioritize preserving visually pleasing details but focuses instead on retaining features essential for specific machine tasks. For example, in vehicle counting systems that rely on motion features, compression can selectively preserve important features while discarding irrelevant textures or details. Such task-specific compression reduces the bitrate significantly without degrading analytical performance. 

\begin{figure}[t]
    \centering
    \begin{minipage}[b]{1\linewidth}
    \centering
    \includegraphics[width=1\textwidth]{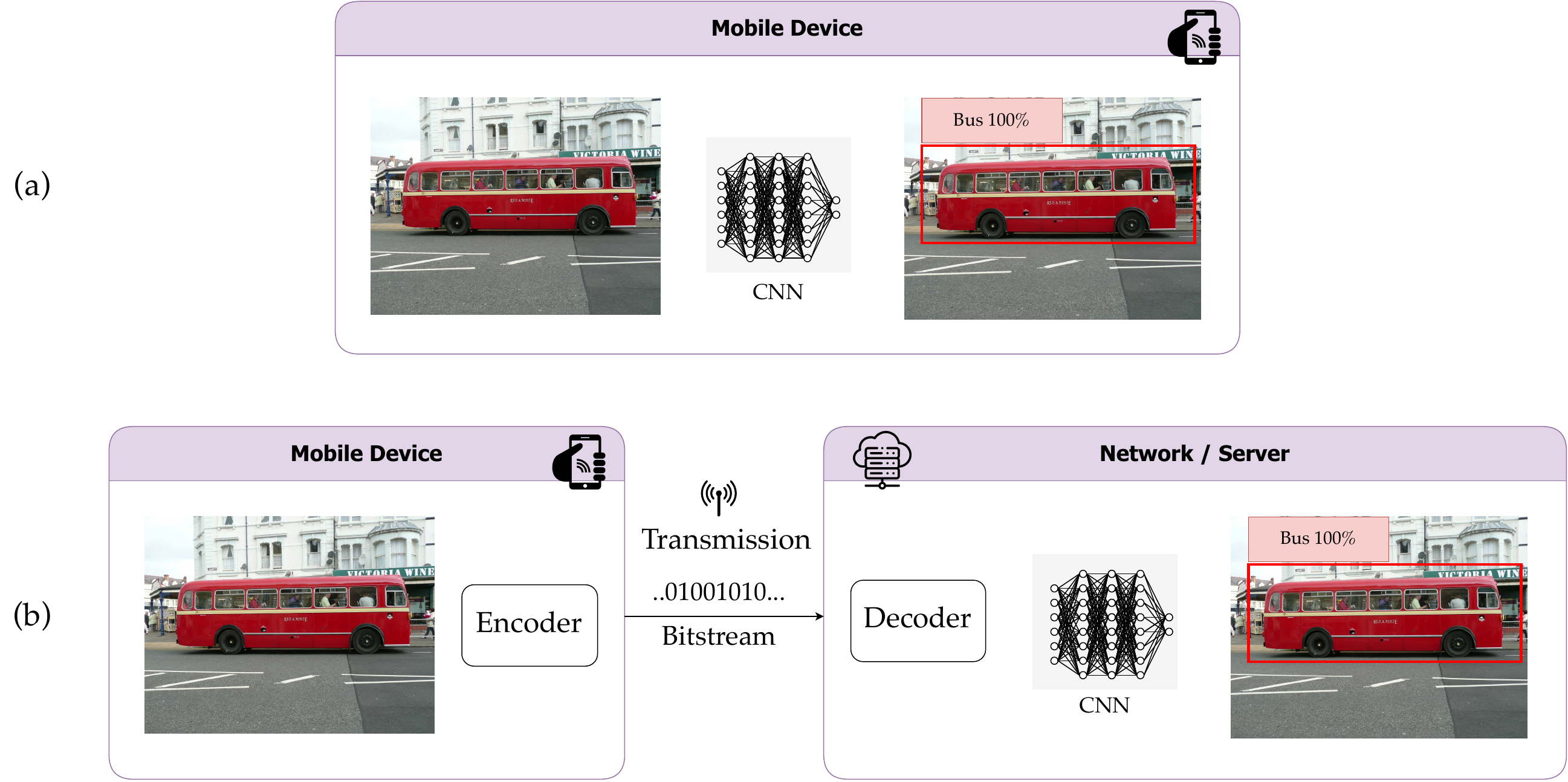}
    \end{minipage}
\vspace{-0.4cm}
\caption{(a) Edge Inference (b) Remote Inference}
\vspace{-0.4cm}
\label{fig:intro}
\end{figure}

Edge inference and remote inference represent two distinct approaches for deploying machine vision models, as shown in Fig.~\ref{fig:intro}. Edge inference runs deep learning models locally on the edge device to generate inference predictions. This method works well for simpler tasks, but the limited computational resources of edge devices make it impractical to run large deep learning models. Deep learning models are progressively incorporating advanced architectures that require substantial memory and processing power. These constraints fundamentally limit the capabilities of edge inference for high-performance applications.

Remote inference, on the other hand, involves transmitting compressed data to a central server with extensive computational resources. This approach allows for the use of computationally intensive and accurate models, leveraging state-of-the-art deep learning architectures to generate faster and more reliable inferences. Furthermore, remote inference facilitates centralized model updates and maintenance, enabling systems to consistently adopt the latest advancements in AI. As deep learning models continue to grow in complexity, optimizing remote inference systems becomes crucial for maintaining scalability and efficiency in machine vision systems. Effective compression tailored for machine consumption ensures that only task-relevant features are transmitted, reducing bandwidth demands while preserving the fidelity required for robust analysis. This optimization is critical to addressing the increasing computational demands of modern AI systems in resource-intensive applications.

\begin{figure*}[t]
    \centering
    \begin{minipage}[b]{1\linewidth}
    \centering
    \includegraphics[width=1\textwidth]{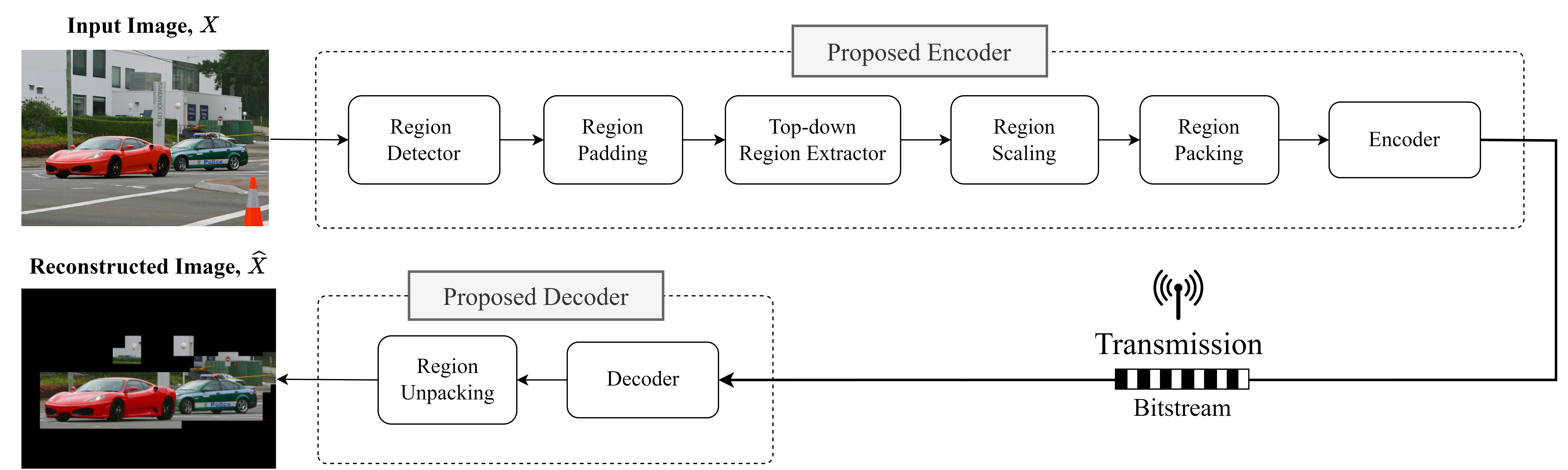}
    \end{minipage}
\caption{Overview of the Proposed Method}
\vspace{-0.5cm}
\label{fig:proposed_method}
\end{figure*}

Recognizing the growing importance and demand for compression techniques tailored for machine vision, the Moving Picture Experts Group (MPEG) initiated efforts to address this emerging need. In July 2019, an Ad-hoc Group was established to investigate the necessity of a dedicated standard and define relevant use cases. This initiative marked the beginning of a structured approach toward creating standardized solutions for machine vision applications. At the 133\textsuperscript{rd} MPEG meeting in July 2021, significant progress was made with the formal development of use case requirements~\cite{vcm_use_case} and an evaluation framework~\cite{vcm_eval}. This framework was decided by extensive input from industry experts, reflecting a collaborative effort to align the standard with real-world demands and diverse application scenarios. Subsequently, at the 138\textsuperscript{th} MPEG meeting in July 2022, a Call for Proposals (CfP) was issued~\cite{vcm_cfp}, officially launching the standard development process. The CfP resulted in baseline proposals from a diverse range of contributors~\cite{vcm_cfp_report}, including Alibaba, the Institute of Computing Technology at the Chinese Academy of Sciences (CAS-ICT), China Telecom, City University of Hong Kong, Ericsson, the Electronics and Telecommunications Research Institute (ETRI), Konkuk University, Myongji University, Nokia, Poznan University of Technology (PUT), Tencent, V-Nova, Wuhan University, Zhejiang University, Florida Atlantic University (FAU), and OP Solutions LLC.

In this paper, we propose a novel approach to image compression specifically designed for machine vision tasks. Unlike traditional approaches, our method discards regions of images that are not critical for end-task accuracy, without the need for retraining or fine-tuning of the deep learning model. The important regions of interest are efficiently packed into a smaller resolution, enabling a superior trade-off between bitrate and accuracy. We have used MPEG's baseline remote inference anchor~\cite{vcm_cttc}, as depicted in Fig.~\ref{fig:intro}(b), with Versatile Video Coding (VVC)~\cite{vvc} employed as the encoder and decoder to compare our results. Experimental results demonstrate that the proposed approach achieves a bitrate reduction of up to 44.10\% without sacrificing the end-task accuracy. At the same bitrate, our proposed method improves the end-task accuracy by up to 8.88\% compared to MPEG's remote inference anchor.

The remainder of this paper is organized as follows: Section II presents the proposed method in detail. Section III describes the experimental setup and provides a comprehensive analysis of the results. Finally, Section IV concludes the paper.

\section{Proposed Method}

Our proposed method for compressing images for machine vision focuses on important regions of interest in images to reduce the bitrate without sacrificing the end-task accuracy of the system.   Overview of the proposed method is depicted in Fig.~\ref{fig:proposed_method}. The input image, \(X\), is processed by our proposed encoder, producing a bitstream that is subsequently decoded by the decoder. The reconstructed image, \(\hat{X}\), is then passed through the end-task network. Our proposed encoder consists of six modules i.e., Region Detector, Region Padding, Top-down Region Extractor, Region Scaling, Region Packing, and Encoder. However, our proposed decoder is simple and only consists of two modules i.e., Decoder and Region Unpacking. All the modules are described in detail in the subsequent sections.

\subsection{Region Detector}
The region detector module is designed to identify and output bounding boxes that define key regions of interest within an image. These regions are deemed important for machine vision tasks, as they contain different objects of interest. In scenarios where multiple regions of interest are detected, the bounding boxes may overlap, representing areas of shared significance. Depending on the content of the input frame, the detected regions may vary significantly. For instance, in some cases, the entire area of the frame may be covered by the detected regions, while in others, only a small portion of the frame may be included. 

Although any deep learning model can be used as the region detector, the complexity of this module should remain below that of the deep learning model used for the end-task to maintain computational efficiency. In our implementation, we used YOLOv7~\cite{yolov7} for its fast inference speed and high detection accuracy. Despite the choice of YOLOv7 in this paper, the region detector can be replaced with simple, low-complexity region detection on the edge device. The use of a similar deep learning architecture for both region detection and the end-task offers the potential for maximum performance consistency and better bitrate-accuracy trade-off as it will be able to precisely identify the regions of interest for the end-task network. 


\subsection{Region Padding}

Identified regions of interest can be extended to provide additional context for each detection, contributing to improved endpoint machine task performance. This extension introduces a padding mechanism for inference objects, ensuring that the boundaries of the detected regions are sufficiently expanded to capture contextual information that may enhance downstream tasks. In the proposed implementation, an inference boundary extension of 15 pixels is applied to each detection. By expanding the boundaries, the module aims to incorporate peripheral details that may otherwise be excluded, leading to superior end-task performance.
\begin{figure}[t]
    \centering
    \begin{minipage}[b]{1\linewidth}
    \centering
    \includegraphics[width=1\textwidth]{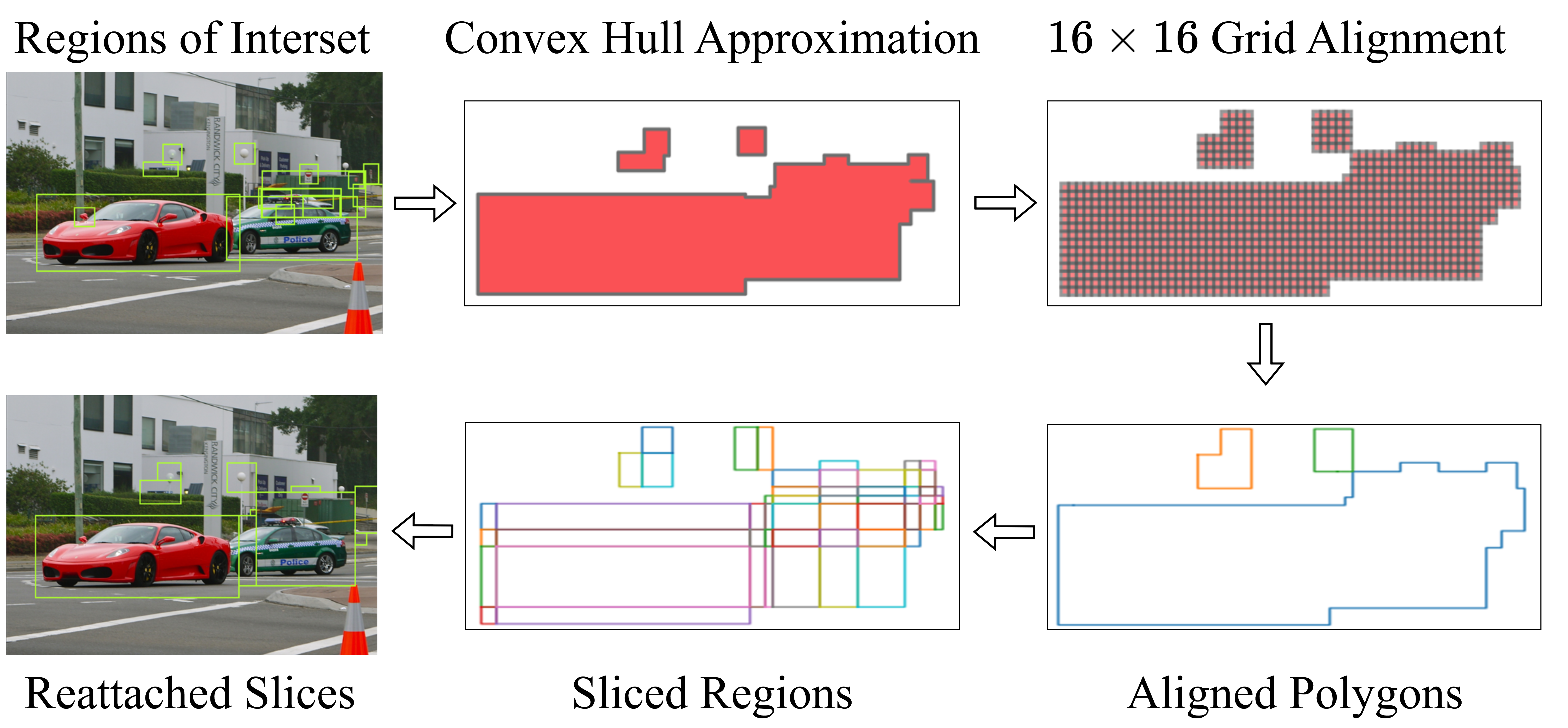}
    \end{minipage}
\caption{Top-down Region Extractor}
\label{fig:top-down-roi-extract}
\vspace{-0.04in}
\end{figure}

\subsection{Top-down Region Extractor}

This module merges overlapping regions to generate the smallest possible convex polygon that encloses all the regions of interest using a \emph{Convex Hull}. Given a set of \( n \) regions \( R_i \), where each region \( R_i \) is defined by its top-left corner \((x_i, y_i)\), width \(w_i\), and height \(h_i\), and each region \( R_i \) is represented by its four vertices:
\begin{align}
    v_1^i &= (x_i, y_i), \quad \\
    v_2^i &= (x_i + w_i, y_i), \quad  \\
    v_3^i &= (x_i, y_i + h_i), \quad  \\
    v_4^i &= (x_i + w_i, y_i + h_i), \quad  
\end{align}

The set of all vertices across the \( n \) regions could be defined as:
\begin{equation}
    P = \bigcup_{i=1}^n \{v_1^i, v_2^i, v_3^i, v_4^i\}.
\end{equation}

The Convex Hull \( CH(P) \) of the vertex set \( P \) is the smallest convex polygon that contains all points in \( P \). It can be formally expressed as:
\begin{equation}
    CH(P) = \text{Convex Hull} \left( \{p_1, p_2, \dots, p_m\} \right),
\end{equation}
where \( \{p_1, p_2, \dots, p_m\} \subseteq P \) are the vertices of the Convex Hull.

Then, a \(16\times16\) grid structure is employed to pad and extend the unified regions , \(CH(P)\). This grid-based alignment ensures that regions will be aligned with the coding units of the encoder to achieve better compression efficiency.  After the alignment, the resulting extended polygon is further processed by splitting them into rectangular sub-boxes. This is achieved by slicing the input polygons based on their non-rectangular vertices. To enhance packing efficiency, the resulting slices are greedily and recursively reassembled based on shared edges, resulting in improved rectangular box configurations as showed in Fig.~\ref{fig:top-down-roi-extract}. 

\subsection{Region Scaling}
In this module, certain regions of the images are adaptively down-scaled to enhance compression efficiency while reducing the bitrate. Our experiments indicate that maintaining the original resolution for all the regions of interest is not necessary. For instance, regions of interest primarily containing cars or person can undergo down-scaling without negatively affecting the performance of end-task accuracy. 

\begin{figure}[t]
    \centering
    \begin{minipage}[b]{1\linewidth}
    \centering
    \includegraphics[width=1\textwidth]{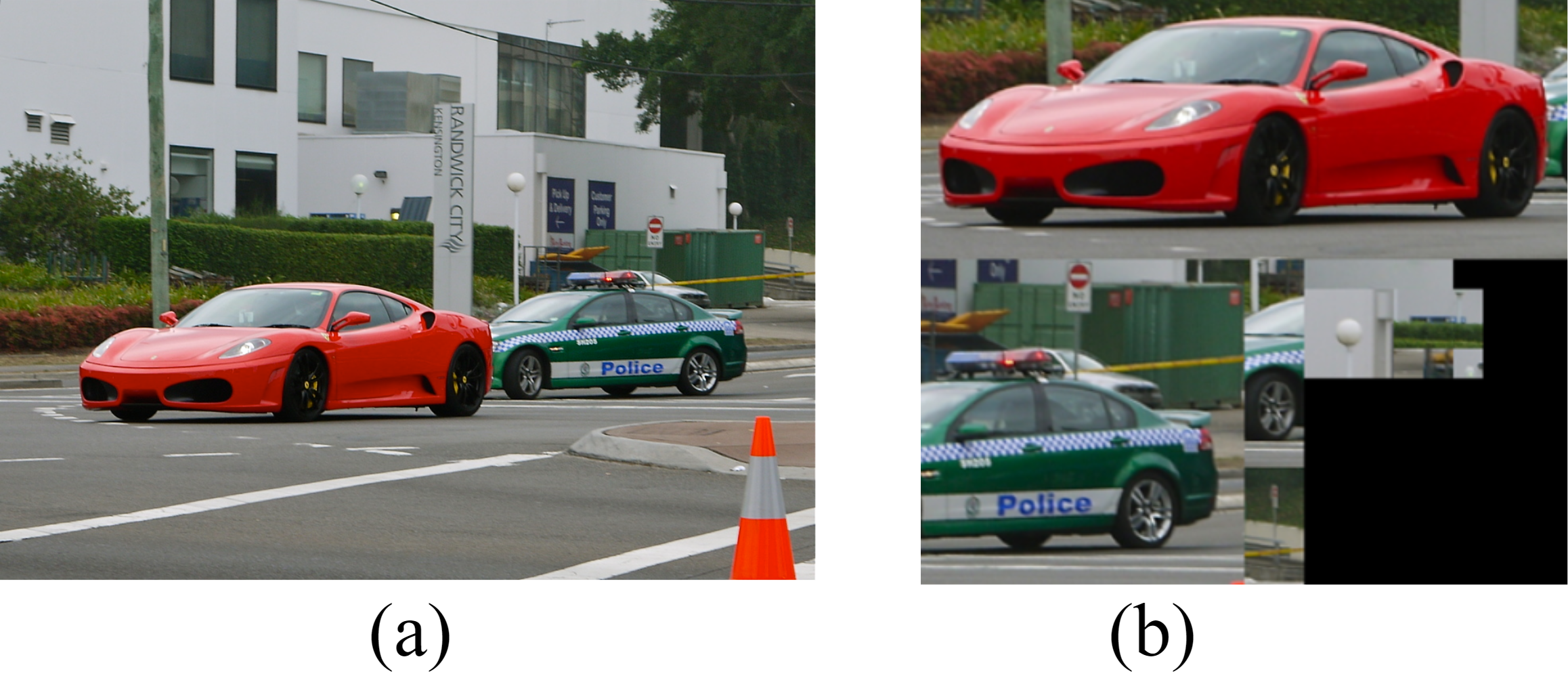}
    \end{minipage}

\caption{(a) Original Image (Resolution: \(1024\times730\)) (b) Packed Image (Resolution: \(352\times330\)))}
\label{fig:packed_image}
\vspace{-0.1in}
\end{figure}

\begin{table*}[t]
\centering
\renewcommand{\arraystretch}{1.2}
\setlength{\tabcolsep}{8pt}
\caption{BD-Rate and BD-mAP for Remote Inference VS Proposed Method}
\label{tab:bd_rate_map}
\begin{tabular}{|l|l|l|c|ccc|}
\toprule
\textbf{Task} & \textbf{Network} & \textbf{Dataset} & \textbf{Image Type} & \textbf{Size (\%)} & \textbf{BD-rate (\%)} & \textbf{BD-mAP (\%)} \\ 
\midrule
\multirow{12}{*}{\textbf{Object Detection}} 
    & \multirow{12}{*}{Faster RCNN-X101-FPN~\cite{faster_rcnn}} 
    & \multirow{4}{*}{FLIR~\cite{flir_thermal_dataset}}        
    & Infrared & 100 & -10.86 & 0.60 \\ 
    & & & Infrared & 75  & -18.10 & 2.17 \\ 
    & & & Infrared & 50  & -18.77 & 2.52 \\ 
    & & & Infrared & 25  & -7.21  & 0.47 \\ \cmidrule(lr){3-7}
    & & \multirow{4}{*}{OpenImages~\cite{oiv6_det}} 
    & RGB      & 100 & -19.93 & 2.02 \\ 
    & & & RGB      & 75  & -17.38 & 2.06 \\ 
    & & & RGB      & 50  & -13.67 & 1.93 \\ 
    & & & RGB      & 25  & -3.52  & 0.34 \\ \cmidrule(lr){3-7}
    & & \multirow{4}{*}{TVD~\cite{tvd}}         
    & RGB      & 100 & -40.03 & 7.26 \\ 
    & & & RGB      & 75  & -41.94 & \textbf{8.88} \\ 
    & & & RGB      & 50  & -39.13 & 8.60 \\ 
    & & & RGB      & 25  & -20.95 & 3.09 \\ 
\midrule
\multirow{8}{*}{\textbf{Instance Segmentation}} 
    & \multirow{8}{*}{Mask RCNN-X101-FPN~\cite{mask_rcnn}}  
    & \multirow{4}{*}{OpenImages~\cite{oiv6_seg}} 
    & RGB      & 100 & -31.00 & 3.50 \\ 
    & & & RGB      & 75  & -22.48 & 2.98 \\ 
    & & & RGB      & 50  & -17.93 & 2.80 \\ 
    & & & RGB      & 25  & -6.62  & 1.16 \\ \cmidrule(lr){3-7}
    & & \multirow{4}{*}{TVD~\cite{tvd}}        
    & RGB      & 100 & -42.47 & 5.71 \\ 
    & & & RGB      & 75  & \textbf{-44.10} & 7.18 \\ 
    & & & RGB      & 50  & -39.60 & 6.65 \\ 
    & & & RGB      & 25  & -21.30 & 2.66 \\ 
\bottomrule
\end{tabular}
\end{table*}

\subsection{Region Packing}
The extracted transformed regions get efficiently consolidated into a single frame through the application of a bin packing algorithm in this module. Bin packing~\cite{bin_packing} algorithms aim to arrange a set of items (in this case, image regions) of varying sizes into a finite number of bins (or a single target frame) with fixed dimensions while minimizing unused space. 

Let there be a set of \( n \) regions \( R_i = (w_i, h_i) \), where \( w_i \) and \( h_i \) denote the width and height of region \( R_i \), into a single target frame of fixed dimensions \( W_{\text{bin}} \times H_{\text{bin}} \). The placement of each region \( R_i \) is represented by its top-left corner coordinates \( (x_i, y_i) \), subject to the following boundary constraints:
\begin{align}
    x_i + w_i &\leq W_{\text{bin}}, \label{eq:boundary_x} \\
    y_i + h_i &\leq H_{\text{bin}}. \label{eq:boundary_y}
\end{align}

At each iteration, the algorithm evaluates all the non-overlapping feasible positions for \( R_i \) and selects the one that minimizes the remaining unused space:
\begin{equation}
    S_{\text{unused}} = W_{\text{bin}} \times H_{\text{bin}} - \sum_{j=1}^k (w_j \times h_j), \label{eq:unused_space}
\end{equation}
where \( k \) is the number of regions already placed. By iteratively selecting the best-fit positions for all regions, the method achieves tight packing, reducing spatial gaps and optimizing the overall utilization of the target frame. All the empty spaces of the target frame are colored with black pixels at the end of the bin packing. Additionally, this method ensures that the regions of interest, even when down-scaled, are organized into a single coherent output, facilitating effective encoding. An example of a packed image is shown in Fig.~\ref{fig:packed_image}. It should be noted that in this example, the image resolution after packing is  \(352\times330\), whereas the original image resolution was \(1024\times730\). This method efficiently reduce the image resolution  by discarding the unimportant regions of the image, which consequently provide high compression efficiency without degrading the performance of the end-task network.

\subsection{Encoder}
After the regions are adaptively scaled and tightly packed into a single frame, the resulting image is converted to the YUV format, a widely-used chroma sub-sampling method for efficient compression~\cite{yuv}. The packed image is then compressed using the Versatile Video Coding (VVC)~\cite{vvc} test model (VTM)~\cite{vtm}, which represents the state-of-the-art in video compression.  VVC was chosen for this work because it offers significant improvements over previous standards like H.265/HEVC~\cite{hevc} and H.264/AVC~\cite{avc}. These improvements include advanced block partitioning, improved transform coding, and sophisticated entropy coding, which enable high-quality compression at lower bitrates. 

The VVC codec was configured in All-Intra mode and the necessary region parameters describing the spatial positions, dimensions, and scaling factors of the individual regions are multiplexed into the bitstream. This metadata allows the decoder to accurately unpack the compressed frame and reconstruct all regions in their original positions. 
 
\subsection{Decoder}
At the decoder side, the compressed bitstream is first decoded using VTM. The use of VTM ensures high-quality reconstruction of the packed frame with minimal loss, aligning with the stringent requirements of downstream tasks. In addition to the packed frame, the bitstream also contains multiplexed metadata describing the spatial positions, dimensions, and scaling factors of the individual regions.


\subsection{Region Unpacking}

The Region Unpacking process begins by parsing the metadata to identify the location and size of each region within the packed frame. Each region is then extracted and, if it was scaled during the encoding process, resized back to its original dimensions using the scaling factors provided in the metadata. The regions are subsequently reassembled into their original positions within the reconstructed image, ensuring that the spatial alignment matches the pre-encoding layout. After repositioning all the regions to their original position, empty places are set to black pixels and then the resulting image is passed to the end-task network.

\section{Experimental Result}
\begin{figure*}[t]
    \centering
    \begin{minipage}[b]{1\linewidth}
    \centering
    \includegraphics[width=1\textwidth]{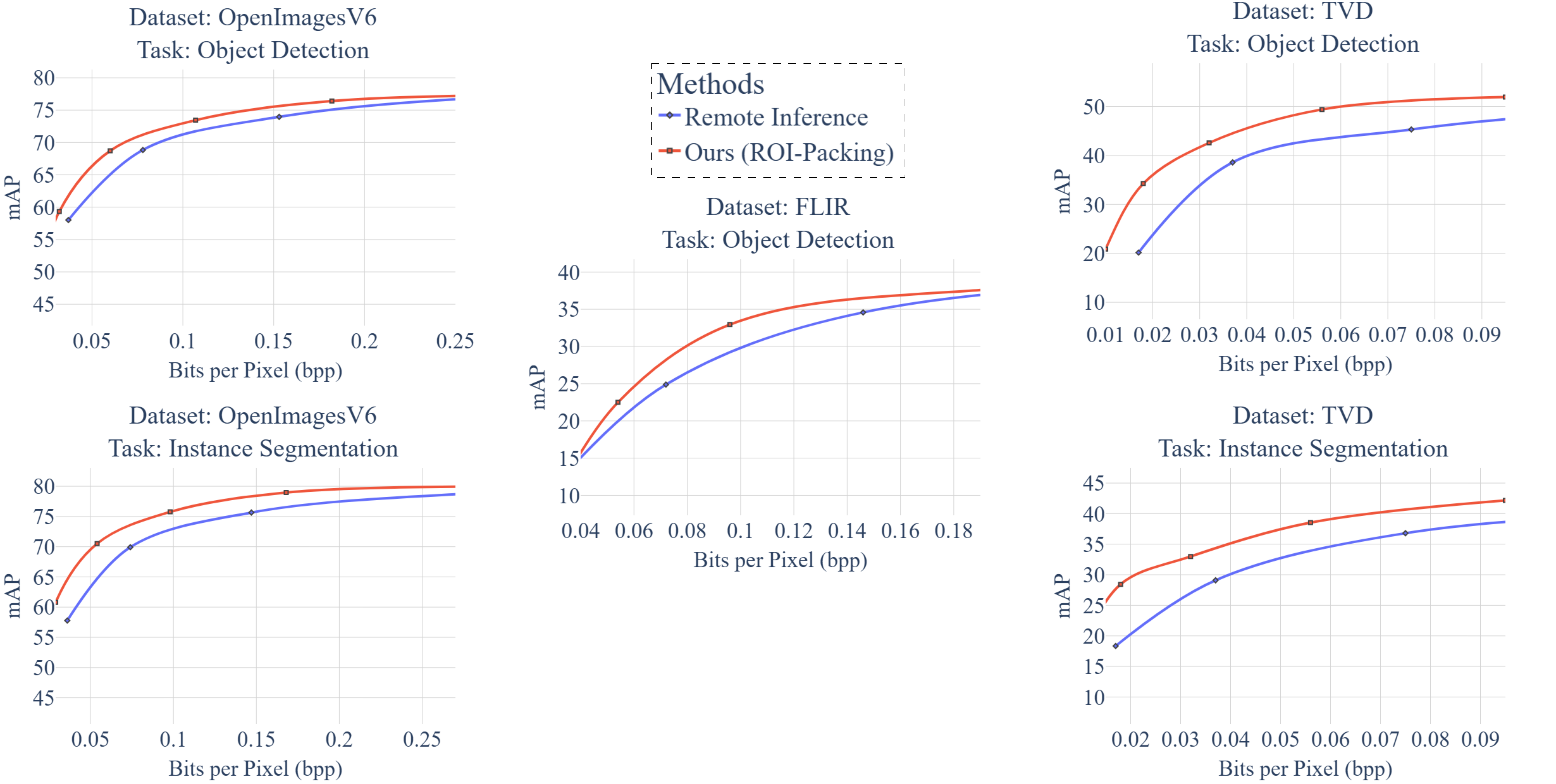}
    \end{minipage}

\caption{Rate-Accuracy Plots for Object Detection \& Instance Segmentation }
\label{fig:rd_plots}
\vspace{-0.1in}
\end{figure*}

\subsection{Experimental Setup}

In this study, the Common Test Conditions (CTC)~\cite{vcm_cttc} and evaluation framework~\cite{vcm_eval} provided by MPEG were used to generate the remote inference anchor and evaluate the proposed method. It is noteworthy that the CTC was established through consensus among members of the MPEG expert group~\cite{vcm_use_case}, reflecting a broad alignment of interests from diverse industry use cases. To ensure a comprehensive evaluation, two distinct types of image datasets—infrared and RGB—were employed across five datasets specified in the CTC. These datasets were selected to validate the proposed method's capability to generalize effectively across varying input types and challenging scenarios. The evaluation was conducted on two end-tasks: object detection and instance segmentation, chosen for their relevance in real-world applications such as surveillance and autonomous systems.

For the object detection task, the Faster R-CNN-X101-FPN~\cite{faster_rcnn} was employed. This network is well-regarded for its robust region proposal mechanism and efficient feature pyramid structure, which enables accurate object detection across multiple scales. For the instance segmentation task, the Mask R-CNN-X101-FPN~\cite{mask_rcnn} was utilized. This network extends Faster R-CNN by incorporating a dedicated branch for predicting segmentation masks by enabling pixel-level classification in addition to bounding box detection. Both networks were built with the ResNeXt-101~\cite{resnet} backbone and Feature Pyramid Networks (FPN)~\cite{fpn}, facilitating enhanced feature extraction across diverse spatial resolutions and improving performance on complex datasets.

Additionally, the next generation VVC standard test model, commonly referred to as VTM~\cite{vtm}, was employed as both the encoder and decoder using the ``All-Intra" configuration and ``Still Picture Profile''. To evaluate the performance of the proposed method under varying compression levels, six Quantization Parameters (QPs), i.e., \(22\), \(27\), \(32\), \(37\), \(42\), and \(47\), were selected. These QPs represent a wide range of bitrate-accuracy trade-offs, ensuring that the evaluation accounts for diverse operational scenarios. Furthermore, the proposed approach's efficacy was assessed at four different input sizes, specifically scaled at 100\%, 75\%, 50\%, and 25\%, to simulate various resource-constrained conditions. This comprehensive evaluation framework highlights the robustness and scalability of our proposed method in practical scenarios.

\subsection{Result Analysis}

The performance of the proposed method in comparison to MPEG's remote inference anchor is presented in Table~\ref{tab:bd_rate_map} in terms of BD-Rate and BD-mAP. BD-Rate is a widely used metric for evaluating the bitrate savings between two different codecs~\cite{bd_rate}.

As shown in Table~\ref{tab:bd_rate_map}, the BD-Rate for the proposed method is consistently negative, implying that it always requires fewer bits than the remote inference anchor. Remarkably, the proposed method achieves a potential bitrate reduction of up to \(44.10\%\). This significant reduction demonstrates that the proposed approach can deliver the same end-task accuracy as the remote inference anchor while using 44.10\% fewer bits, enhancing system efficiency. Both the object detection and instance segmentation tasks exhibit substantial bitrate savings with the proposed method compared to the anchor. Additionally, it is observed that the BD-Rate is lower at size 25\% in most of the cases. This outcome is expected, as size 25\% images are down-scaled by a factor of 0.25 of original image resolution and contain fewer extraneous regions that can be discarded.

In addition to BD-Rate, BD-mAP~\cite{vcm_eval} evaluates the impact of a codec on the mean Average Precision (mAP) at a fixed bitrate. A positive BD-mAP value indicates that a codec achieves higher mAP compared to its counterpart at the same bitrate, signifying improved performance. Inversely, a negative BD-mAP value suggests that the codec delivers lower mAP than its counterpart under identical bitrate conditions. As shown in Table~\ref{tab:bd_rate_map}, the BD-mAP of the proposed method is consistently positive, indicating that it consistently provides higher end-task accuracy at the same bitrate. Notably, the proposed method shows a potential BD-mAP improvement of up to \(8.88\%\), highlighting its ability to significantly improve end-task performance without requiring additional bits. These results unequivocally underscore the effectiveness of the proposed method in optimizing bitrate efficiency and improving task accuracy across both object detection and instance segmentation tasks.

Furthermore, the rate-accuracy plots for all datasets are presented in Fig.~\ref{fig:rd_plots}. For brevity, only plots for the 100\% size configuration are shown. In these plots, the x-axis represents bits per pixel (bpp), while the y-axis indicates the end-task accuracy measured in mAP. The red curve corresponds to the proposed method, and the blue curve represents MPEG's remote inference anchor.  It is evident from the plots that, in all cases, the proposed method achieves higher end-task accuracy compared to the anchor. However, it is worth noting that for the FLIR dataset, which contains infrared images, performance at the lowest bpp point is comparable to the anchor. This suggests that the proposed method is sensitive to infrared images at low quality. Despite this limitation, the proposed method consistently demonstrates superior accuracy performance across all other datasets \& tasks.

\section{Conclusion}
ROI-Packing introduces a novel paradigm in image compression tailored for machine vision, achieving significant advancements in both bitrate efficiency and task accuracy. The proposed method demonstrates substantial performance improvements, including up to 44.10\% bitrate savings and an 8.88\% accuracy enhancement over state-of-the-art codecs. Furthermore, the method's compatibility with existing end-task models, without the need for retraining or fine-tuning, establishes it as a versatile and implementable solution for various machine vision applications in practical scenarios. Future work could extend ROI-Packing to video compression by applying it to every intra-frame period, potentially enhancing performance in continuous video streams. 

\newpage

\bibliographystyle{IEEEbib}
\bibliography{icme2025references}

\end{document}